\documentclass[]{mic2011}

\usepackage{cite}
\usepackage[ruled,vlined,commentsnumbered]{algorithm2e}
\usepackage{url}
\usepackage{pstricks}
\usepackage{pstricks-add}

\begin{document}

\title{On the use of reference points for the biobjective\\Inventory Routing Problem}
\author{M. J. Geiger\inst{1} \and M. Sevaux\inst{1,2}}
\institute{
  Helmut Schmidt University\\
    Logistics Management Department\\
    Holstenhofweg 85, 22043 Hamburg, Germany\\
  \email{m.j.geiger@hsu-hh.de}
  \and
  Universit\'e de Bretagne-Sud\\
  Lab-STICC -- Centre de Recherche\\
  2 rue de St Maud\'e, F-56321 Lorient, France\\
  \email{marc.sevaux@univ-ubs.fr}
}
\id{S1-14}
\maketitle

\begin{abstract}
The article presents a study on the biobjective inventory routing problem. Contrary to most previous research, the problem is treated as a true multi-objective optimization problem, with the goal of identifying Pareto-optimal solutions. Due to the hardness of the problem at hand, a reference point based optimization approach is presented and implemented into an optimization and decision support system, which allows for the computation of a true subset of the optimal outcomes. Experimental investigation involving local search metaheuristics are conducted on benchmark data, and numerical results are reported and analyzed.
\end{abstract}

\section{Introduction}
The Inventory Routing Problems (IRP) is a true multi-objective optimization problem, in which decisions from inventory management intersect with the classical vehicle routing problem (VRP). In its context, delivery quantity decisions need to be made, which then lead to a series of multi-period vehicle routing problems. Several different capacity constraints, e.\,g.\ at the customers' and of the trucks, influence the choice of the delivery quantities. Consequently, the two areas overlap to a considerable degree, and thus the solution of the IRP is not only $\mathcal{NP}$-hard but also challenging when `only' trying to find good solutions by means of heuristic approaches.

A number of articles on the IRP has appeared in the past. While first approaches can be traced back to the 1980s~\cite{bell.dalberto.ea:83,federgruen:1984}, a considerable growth becomes apparent in past years, with a recent survey given in~\cite{bertazzi:2008:incollection}. Often, metaheuristics are used to solve the IRP. Some prominent examples are Iterated Local Search \cite{ribeiro:2003:techreport}, Variable Neighborhood Search \cite{zhao:2008:article}, Greedy Randomized Adaptive Search \cite{campbell:2004:article:2}, and Memetic Algorithms \cite{boudia:2009:article}.

Common to most previous work on the problem is the combination of the two aspects, i.\,e.\ the minimization of the inventory levels and the routing costs, into an overall cost function. This is possible if such partial cost functions can be formulated. Then however, the tradeoff between the two aspects is not delivered in those mono-criterion models. When planning on a tactical level however, such issues may quickly arise. Here, the impact of changes in one partial cost-function on the other becomes relevant. Practical examples include changes in fuel prices, the orientation towards `green' logistics, and similar considerations. We believe that multi-objective models may contribute to the better understanding of the IRP with respect to such tactical planning situations, and consequently present a study on this topic.

In the following Section~\ref{sec:problem:definition}, we briefly introduce the investigated IRP. The methodological approach used to solve the problem is then described in Section~\ref{sec:methodological:approach}. Experiments and results on available benchmark data follow in Section~\ref{sec:experiments}, and conclusions are presented in Section~\ref{sec:conclusions}.

\section{\label{sec:problem:definition}Description of the investigated IRP}
In line with our formulation described in \cite{geiger:2011}, we consider an IRP in which a given set of $n$ customers needs to be delivered with goods from a single depot. With respect to the vehicle routing part of our IRP, the characteristics of the classical capacitated vehicle routing problem (CVRP) are considered. This means that we assume a unlimited number of available trucks, each of which has a limited capacity $C$ for the delivered goods.

Decision variables of the problem are on the one hand delivery quantities $q_{it}$ for each customer $i$, $i = 1, \ldots, n$, and each period $t$, $t = 1, \ldots, T$. On the other hand, a VRP must be solved for each period $t$ of the planning horizon $T$, combining the delivery quantities $q_{it}$ into tours/routes for the involved fleet of vehicles. We assume $q_{it} \geq 0 \,\, \forall i,t$, thus forbidding the pickup of goods at all times. Following the definition of the classical CVRP, we do not permit split-deliveries, and therefore $q_{it} \leq C \,\, \forall i,t$.

At each customer $i$, a demand $d_{it}$ is to be satisfied for each period $t$. Inventory levels $L_{it}$ at the customers are limited to a maximum amount of $Q_{i}$, i.\,e.\ $L_{it} \leq Q_{i} \,\, \forall i, t$. An incoming material flow $\phi^{+}_{it}$ and an outgoing material flow $\phi^{-}_{it}$ links the inventory levels at the customers over the time horizon: $L_{it+1} = L_{it} + \phi^{+}_{it} - \phi^{-}_{it} \,\, \forall i, t =1, \ldots, T-1$. The incoming material flow is given by $\phi^{+}_{it} = q_{it}$ iff $q_{it} \leq Q_{i} - L_{it-1}$, and $\phi^{+}_{it} = Q_{i} - L_{it-1}$ otherwise. On the one hand, this implies that the maximum inventory levels are never exceeded, independent from the choice of the shipping quantities $q_{it}$. On the other hand, delivery quantities $q_{it}$ with $q_{it} > Q_{i} - L_{it-1}$ can be excluded when solving the problem. Obviously, as all inventory levels $L_{it} \geq 0$, any delivery quantity $q_{it} > Q_{i}$ will not play a role also. The outgoing material flow assumes $\phi^{-}_{it} = d_{it}$ iff $d_{it} \leq L_{it-1} + \phi^{+}_{it}$, and $\phi^{-}_{it} = L_{it-1} + \phi^{+}_{it}$ otherwise. The latter case is also referred to a stockout-situation, and the stockout-level can be computed as $d_{it} - L_{it-1} - q_{it}$. As we however assume $d_{it}$ to be known in advance, we can easily avoid such situations by shipping enough goods in advance or just in time.

Two objective functions are considered, leading to a biobjective formulation. First, we minimize the total inventory as given in Expression~(\ref{eqn:inventory}). Besides, the minimization of the total distances traveled by the vehicles for shipping the quantities in each period is of interest. While the solution of this second objective as such presents an $\mathcal{NP}$-hard problem, we denote this second objective with Expression~(\ref{eqn:distance}).

\begin{eqnarray}
  \label{eqn:inventory}\min && \sum_{i=1}^{n} \sum_{t=1}^{T} L_{it}\\
  \label{eqn:distance}\min && \sum_{t=1}^{T} \mbox{VRP}_{t}(q_{1t}, \ldots, q_{nt})
\end{eqnarray}

The two objectives are clearly in conflict to each other. While large shipping quantities allow a minimization of the routes, small quantities lead to low inventory levels over time. In between, we can expect numerous compromise solutions with intermediate values.

Without any additional information or artificial assumptions about the tradeoff between the two functions, no sensible aggregation into an overall evaluation is possible. Consequently the solution of the problem at hand lies in the identification of all optimal solutions in the sense of Pareto-optimality, which constitute the Pareto-set $P$.

\section{\label{sec:methodological:approach}Solution representation and heuristic search for the IRP}
The solution approach of this article is based on the decomposition of the problem into two phases. First, quantity decisions of when and how much to ship to each customer are made. This is done such that only customers running out of stock are served, i.\,e.\ we set $q_{it} > 0$ iff $d_{it} > L_{it-1} \,\, \forall i, t$. Second, the resulting CVRP is solved for each period $t$ by means of an appropriate solution approach.

Solving the IRP in this sequential manner has several advantages. Most importantly, we are able to introduce a compact encoding for the first phase, which allows an intuitive illustration of how the results of the procedure are obtained. We believe this to be an important aspect especially for decision makers from the industry. Besides, searching the encoding is rather easy by means of local search/ metaheuristics.

\subsection{Solution encoding}
Alternatives are encoded by a $n$-dimensional vector $\pi = (\pi_{1}, \ldots, \pi_{n})$ of integers. Each element $\pi_{i}$ of the vector corresponds to a particular customer $i$, and encodes for how many periods the customer is served in a row (covered). In the following, we refer to these values $\pi_{i}$ as a customer delivery \emph{`frequency'}. Precisely, if $d_{it} > L_{it-1}$, then $q_{it}$ may be computed as given in Expression~(\ref{eqn:qit}).

\begin{equation}\label{eqn:qit}
  q_{it} = \min \left\{ \sum_{l=t}^{t-1+\pi_{i}} d_{il} - L_{it-1}, %
  Q_{i}- L_{it-1}, C \right\}
\end{equation}

One the one hand, a day-to-day delivery policy is obtained for values of $\pi_{i} = 1$. On the other hand, large values $\pi_{i}$ ship up to the maximum of the customer/ vehicle capacity, and intermediate values lead to compromise quantities in between the two extremes.

On the basis of the delivery quantities, and as mentioned above, VRPs are obtained for each period which do not differ from what is known in the scientific literature.

\subsection{\label{sec:heuristic:search}Heuristic search}
Throughout the optimization runs, an archive $\widehat{P}$ of nondominated solutions is kept that presents a heuristic approximation to the true Pareto-set $P$. By means of local search, and starting from some initial alternatives, we then aim to get closer to $P$, updating $\widehat{P}$ such that dominated solutions are removed, and newly discovered nondominated ones are added.

In theory, any local search strategy that modifies the frequency values of vector $\pi$ might be used for finding better solutions. Clearly, values $\pi_{i} < 1$ are not possible, and exceptional large values do not make sense either due to the upper bounds on the delivery quantities as given in Expression~(\ref{eqn:qit}).

\subsubsection{Construction procedure}
First, we generate solutions for which the delivery frequency assumes identical values for all customers, starting with 1 and increasing the frequency in steps of 1, up to the point where the alternative cannot be added to $\widehat{P}$ any more. Then, we construct alternatives for which the frequency values are randomly drawn between two values: $j$ and $j+1$, thus mixing the frequency values of the first phase and therefore computing some solutions in between the ones with purely identical frequencies. The so obtained values of $\pi$ are then used to determine the delivery quantities as described above.

The vehicle routing problems are then solved using two alternative algorithms. On the one hand, a classical savings heuristic~\cite{clarke:1964:article} is employed, which allows for a comparable fast construction of the required routes/ tours. On the other hand, the more advanced record-to-record travel algorithm~\cite{li:2007:article} is used. Having two alternative approaches is particularly interesting with respect to the practical use of our system. While the first algorithm will allow for a fast estimation of the routing costs, improved results are possible by means of the second approach, however at the cost of more time-consuming computations.

\subsubsection{Improvement procedure}
The improvement procedure in this article is based on modifying the values within $\pi$, and re-solving the subsequently modified VRPs using the algorithms mentioned above. For any element of $\widehat{P}$, we compute the set of neighboring solutions by modifying the values $\pi_{i}$ by $\pm 1$ while avoiding values of $\pi_{i} < 1$. Therefore the maximum number of neighborhood solutions is $2n$. The search for improved solution naturally terminates in case of not being able to improve any element of $\widehat{P}$. In this sense, the approach implements the concept of a multi-point hillclimber for Pareto-optimization.

\begin{figure}[htbp]
  \begin{center}
    \psset{yunit=.7}
    \begin{pspicture}(-.5,-.5)(10,8.5)
      \rput{90}(-.3,4.25){\sffamily \footnotesize Normalized Routing Cost}
      \rput(5,-.5){\sffamily \footnotesize Normalized Inventory Cost}
      \psaxes[labels=none,ticks=none]{->}(0,0)(10,8.5)
      \psdots[dotstyle=o](1,8)(1.2,7.8)(1.5,7.6)(1.6,7.4)(2.1,7)
             (2.7,6)(3,5.6)(4,5.2)(4.3,4.5)(4.5,3.7)(4.8,3.1)
             (5.3,3)(6,2.6)(7,2.5)(7.8,2.3)(9,1.6)(9.5,1)(8.4,2.1)(5.4,2.9)
             (5.7,2.8)
      \psline[linestyle=dotted](1,8)(1,1)(9.5,1)
      \psdots[dotstyle=square*](1,8)(1,1)(9.5,1)
      \psdots[dotstyle=diamond,dotscale=2](1,2.75)(1,4.5)(1,6.25)%
             (3.125,1)(5.25,1)(7.375,1)
      \psset{arrowscale=1.5,ArrowFill=true}
      \psline[linewidth=1.5pt,ArrowInside=->]{-}(2,3.75)(1,2.75)
      \psline[linewidth=1.5pt,ArrowInside=->]{-}(2,5.5)(1,4.5)
      \psline[linewidth=1.5pt,ArrowInside=->]{-}(2,7.25)(1,6.25)
      \psline[linewidth=1.5pt,ArrowInside=->]{-}(4.125,2)(3.125,1)
      \psline[linewidth=1.5pt,ArrowInside=->]{-}(6.25,2)(5.25,1)
      \psline[linewidth=1.5pt,ArrowInside=->]{-}(8.375,2)(7.375,1)
      \psline[linewidth=1.5pt,ArrowInside=->]{-}(2,9)(1,8)
      \psline[linewidth=1.5pt,ArrowInside=->]{-}(2,2)(1,1)
      \psline[linewidth=1.5pt,ArrowInside=->]{-}(10.5,2)(9.5,1)
      \pscircle[linewidth=1pt](1,8){.25}
      \pscircle[linewidth=1pt](9.5,1){.25}
      \pscircle[linewidth=1pt](1,1){.25}
      \psdots*(8.4,2.1)(7,2.5)(5.3,3)(4.5,3.7)(4,5.2)(2.7,6)(2.1,7)
    \end{pspicture}
  \end{center}
  \caption{The reference point (diamond), directions of search and selected solutions (black dots)}
  \label{fig:refset}
\end{figure}
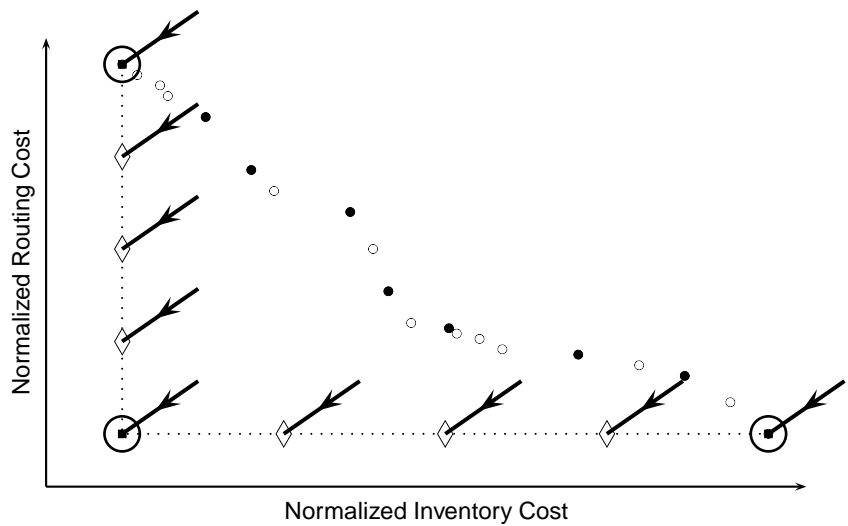

Contrary to searching all elements of $\widehat{P}$, a representative smaller subset of $\widehat{P}$ must be taken for the improvement procedure. This reduces the computational effort, which we have been able to observe in our previous investigations~\cite{geiger:2011}, to a considerable extent. One possibility is to assume a set of reference points, and to select the elements in $\widehat{P}$ that minimize the distances to these reference points. As a general guideline, the reference points should be chosen such that the solutions at the extreme ends of the two objective functions~(\ref{eqn:inventory}) and~(\ref{eqn:distance}) are present in the subset. A favorable implementation of the distance function lies in the use of the weighted Chebychev distance metric, as it allows for a selection of convex-dominated alternatives and thus provides some theoretical advantages over other approaches~\cite{wierzbicki:1999:incollection}. Figure~\ref{fig:refset} illustrates the general use of reference points for the problem at hand. A more detailed discussion of the principle follows in section~\ref{sec:refpoint:selection}.

\vspace{2ex}

\section{\label{sec:experiments}Experimental investigation}
\subsection{Proposition of benchmark data}
Using the geographical information of the 14 benchmark instances given in chapter~11 of \cite{christofides:1979:book}, new data sets for inventory routing have been proposed. While the classical VRP commonly lacks multiple-period demand data, we filled this gap by proposing three demand scenarios, each of which contains a total of $T=240$ demand periods for all customers.

\begin{description}
\item[Scenario a:] The average demand of each customer is constant over time. Actual (integer) demand values for each period however are drawn from an interval of $\pm 25\%$ around this average.
\item[Scenario b:] We assume an increasing average demand, doubling from the initial value at $t=1$ to $t=240$. Again, the actual demand is drawn from an interval of $\pm 25\%$ around this average.
\item[Scenario c:] In this scenario, the average demand doubles from $t=1$ to $t=120$, and goes back to its initial value in $t=240$, following the shape of a sinus curve. Identical to the above presented cases a deviation of $\pm 25\%$ around these values is allowed.
\end{description}

The resulting set of 42 benchmark instances have been made available to the scientific community under \url{http://logistik.hsu-hh.de/IRP} and documented in~\cite{sevaux:2011:techreport}.

\newpage

\subsection{\label{sec:refpoint:selection}Selection of reference points and decision support provided by the method}
\begin{figure}[!ht]
  \centering
  \includegraphics[width=.9\textwidth]{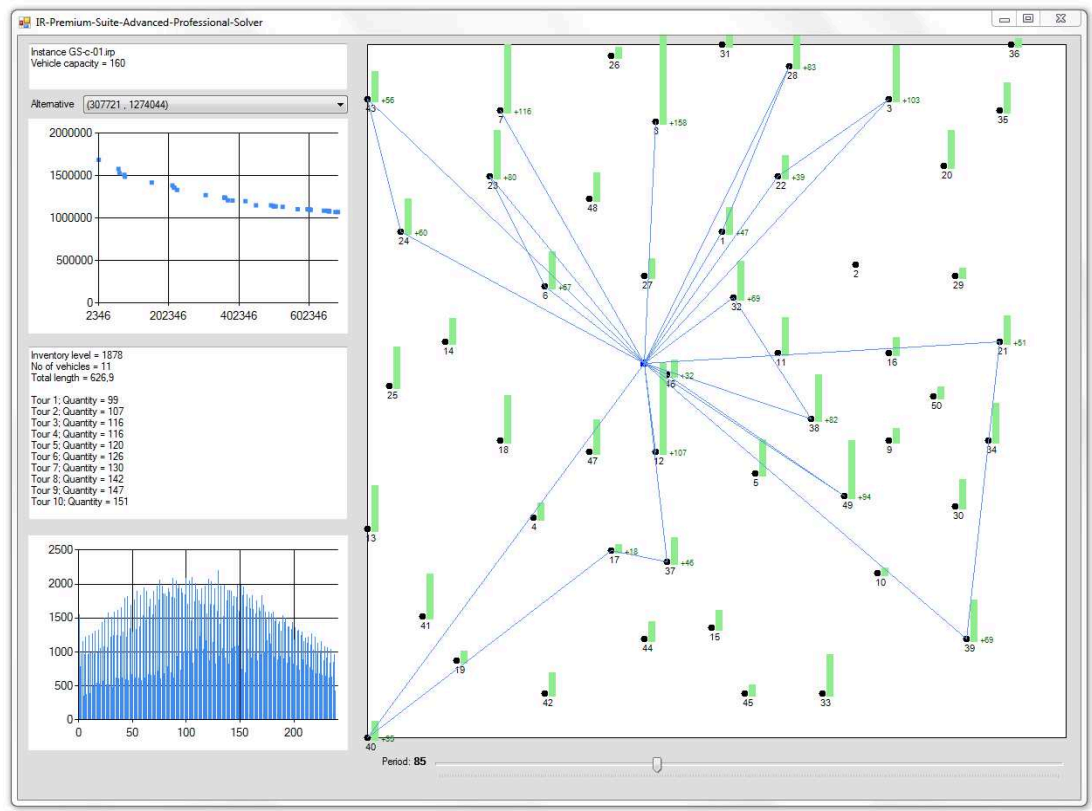}
  \caption{\label{fig:screenshot}Screen shot of the Inventory Routing Solver.}
\end{figure}

A test program has been developed and is used to test new and innovative strategies. Figure \ref{fig:screenshot} shows a typical screen shot of our solver. The upper part on the left gives the name of the instance and the vehicle capacity. Then below, the decision maker is able to display the different alternatives computed by the software.

For the current alternative, and the current period, a text window gives the inventory level, the number of vehicles used and the information on the tours. The box in the bottom left part represents the evolution of the total inventory over all periods. The large window on the right presents the current alternative and period routing. Green bars are the stock level at the customer location for the period chosen with the slide bar at the bottom of the screen.

The subset of alternatives that we select is part of the critical step. We use the above mentioned reference point technique that seems to be the most appropriate in that kind of situations. To make the technique consistent, the two objectives have been normalized. Figure~\ref{fig:refset} depicts a typical situation. The complete set of non-dominated solutions is represented by the black and white dots. The two extreme solutions are highlighted by circles at the top left corner and the bottom right corner. These two extreme solutions plus the ``ideal point'' (the solution circled in the bottom left corner of the figure) having the best value in terms of routing cost and inventory cost compose the three initial reference points.

Then depending on the number of final alternatives that the decision maker wants, we split the two axes (represented by the dotted line) in equal proportions. In the case presented in Figure~\ref{fig:refset}, the two axes have been split in 4, generating 3 additional reference points for the the vertical axe, and 3 additional reference points for the horizontal axe. These new points are represented in the figure by the white diamonds. With each of the 9 reference points, we will select in the complete solution set, the solution that minimizes the Chebyshef distance metric. This will give us at most 9 solutions that will be used in the sequel for improvement as mentioned in section~\ref{sec:heuristic:search}. The direction of search is indicated in Figure~\ref{fig:refset} with the arrows.

\subsection{Results and discussion}
The following Table~\ref{tab:res} reports the computational results of four chosen instances/ scenarios. In total, three different numbers of reference points have been tested: 3, 5, and 11. The choice of the three settings is based on the general explanations given in Figure~\ref{fig:refset}. Three reference points are the minimum possible (sensible) setting for this biobjective case, and represent an approach in which a maximal reduction of the computational effort can be expected. A slightly more refined approximation of the Pareto set can be expected in case of five reference points ($+1$ on each axis). In comparison, the assumption of significantly more (eleven) reference points should then contribute to the better understanding of the relationship of the computational effort to the obtained results.

Obviously, the computational effort needed until converging towards local optima heavily depends on the number of reference points. Besides, the characteristics of the data sets influence the difficulty to a considerable degree. On the one hand, smaller instances are solved faster, which has to be expected, of course. On the other hand, the average demand data has an impact, with instances of scenario type `a' being solved fastest. When comparing scenarios `b' and `c', a clear distinction with respect to the difficulty cannot be made. In some cases, the scenario with the growing average demand turns out to be more demanding, while sometimes the Pareto-optimal solutions of scenario `c' are more costly to approximate.

\begin{table}[!ht]
  \centering
  \resizebox{\textwidth}{!}{
  \begin{tabular}{l|rrrr|rrrr|rrrr|}
    & \multicolumn{4}{c|}{R=3}  & \multicolumn{4}{c|}{R=5} & \multicolumn{4}{c|}{R=11} \\
    Instance & \#Steps & \# eval. & Size & CPU  & \#Steps & \# eval. & Size & CPU  & \#Steps & \# eval. & Size & CPU \\ \hline
    GS-01-a & 14 & 2885 & 113 & 357 & 19 & 5429 & 159 & 800 & 19 & 12660 & 272 & 2344 \\
    GS-01-b &18 & 3339 & 61 & 617 & 18 & 5174 & 85 & 1052 & 19 & 12023 & 187 & 2623 \\
    GS-01-c &27 & 4640 & 78 & 910 & 27 & 6795 & 111 & 1461 & 27 & 18427 & 227 & 4418 \\
    \hline
    GS-02-a &31 & 8780 & 99 & 3295 & 31 & 16123 & 130 & 8950 & 32 & 35430 & 318 & 21753 \\
    GS-02-b &24 & 7438 & 84 & 4827 & 35 & 14086 & 151 & 12255 & 29 & 31035 & 222 & 26768 \\
    GS-02-c &26 & 7532 & 90 & 4024 & 37 & 15891 & 121 & 13000 & 32 & 35087 & 262 & 28909 \\
    \hline
    GS-03-a &45 & 14935 & 204 & 11242 & 48 & 27653 & 334 & 28850 & 46 & 63590 & 663 & 69529 \\
    GS-03-b &49 & 17524 & 130 & 22635 & 49 & 28502 & 253 & 41032 & 49 & 65700 & 522 & 99425 \\
    GS-03-c &62 & 21900 & 57 & 30122 & 62 & 28319 & 98 & 42378 & 65 & 65169 & 373 & 102853 \\
    \hline
    GS-04-a & 66 & 27985 & 302 & 66487 & 66 & 60114 & 495 & 212111 & 67 & 136104 & 791 & 511489 \\
    GS-04-b & 56 & 30381 & 107 & 132146 & 56 & 60050 & 174 & 681293 & 171 & 299372 & 1037 & 1567036 \\
    GS-04-c & 77 & 41539 & 69 & 210539 & 77 & 69768 & 142 & 390263 & 77 & 139568 & 306 & 813263 \\
    \hline
  \end{tabular}
  }
  \caption{Results for the first four instances (all scenarios) and three different numbers of reference points (R=3, R=5, R=11). The number of steps denotes the number of local search steps (neighborhoods) needed to reach a local optimum. Besides, the total number of evaluated alternatives is given, and the size of the obtained approximation $\widehat{P}$ is reported. The computation time (CPU) has been recoded on a single core of an Intel Xeon X5550 processor, and is given in seconds.}
  \label{tab:res}
\end{table}

Overall, it is possible to see that the number of reference points can be used to modulate the computational effort, independent from the particular size or other characteristics of the data sets. With respect to a practical application of our method, this presents a meaningful finding. In an interactive search and decision making approach, it allows for a fast first approximation involving only few reference points, followed by a refined search in a preferred region. Here, the direction of search, i.\,e.\ the preferred region in which additional solutions are to be generated, is given by the decision maker who interacts with the system by means of an appropriate multi-criteria decision aiding technique.

Nevertheless, and especially for the large instances, the computing times are still very high. Clearly, this is due to the assumption of $T = 240$ demand periods. In combination with an increasing number of customers, the computing times grow up to the point where they present a challenge.

\begin{figure}[!ht]
  \centering
    \includegraphics[angle=-90,width=.49\textwidth]{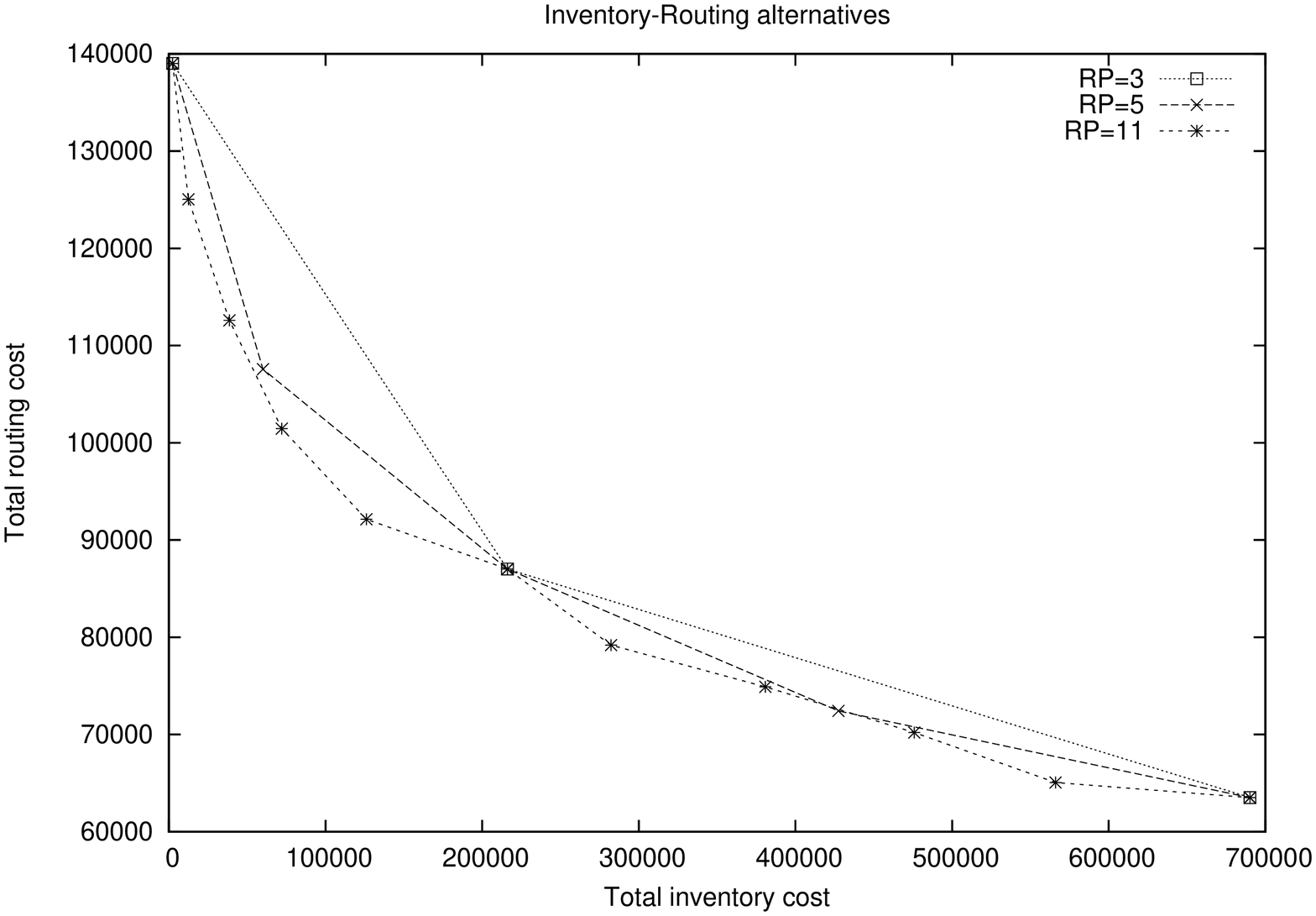}
    \includegraphics[angle=-90,width=.49\textwidth]{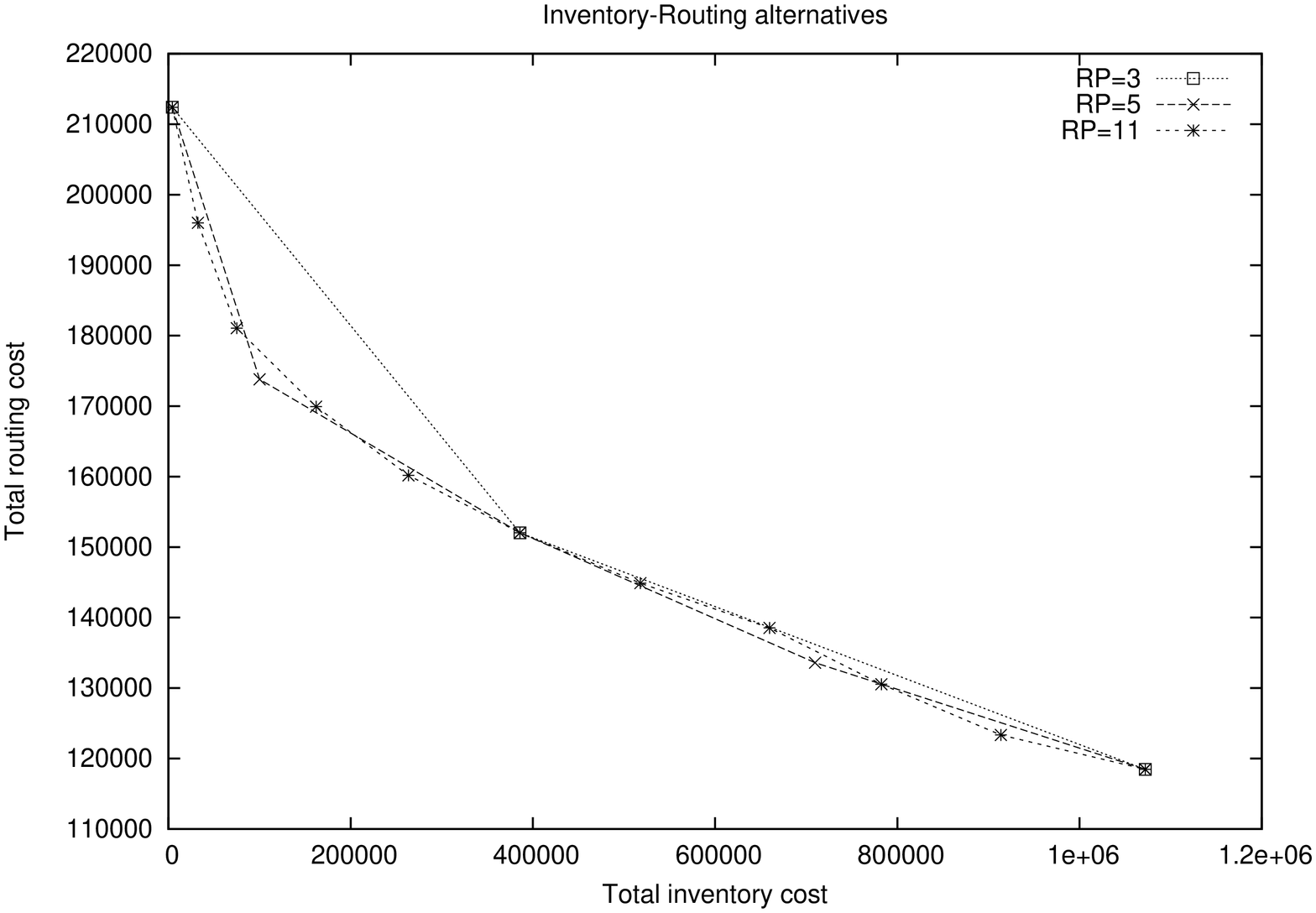}
    \includegraphics[angle=-90,width=.49\textwidth]{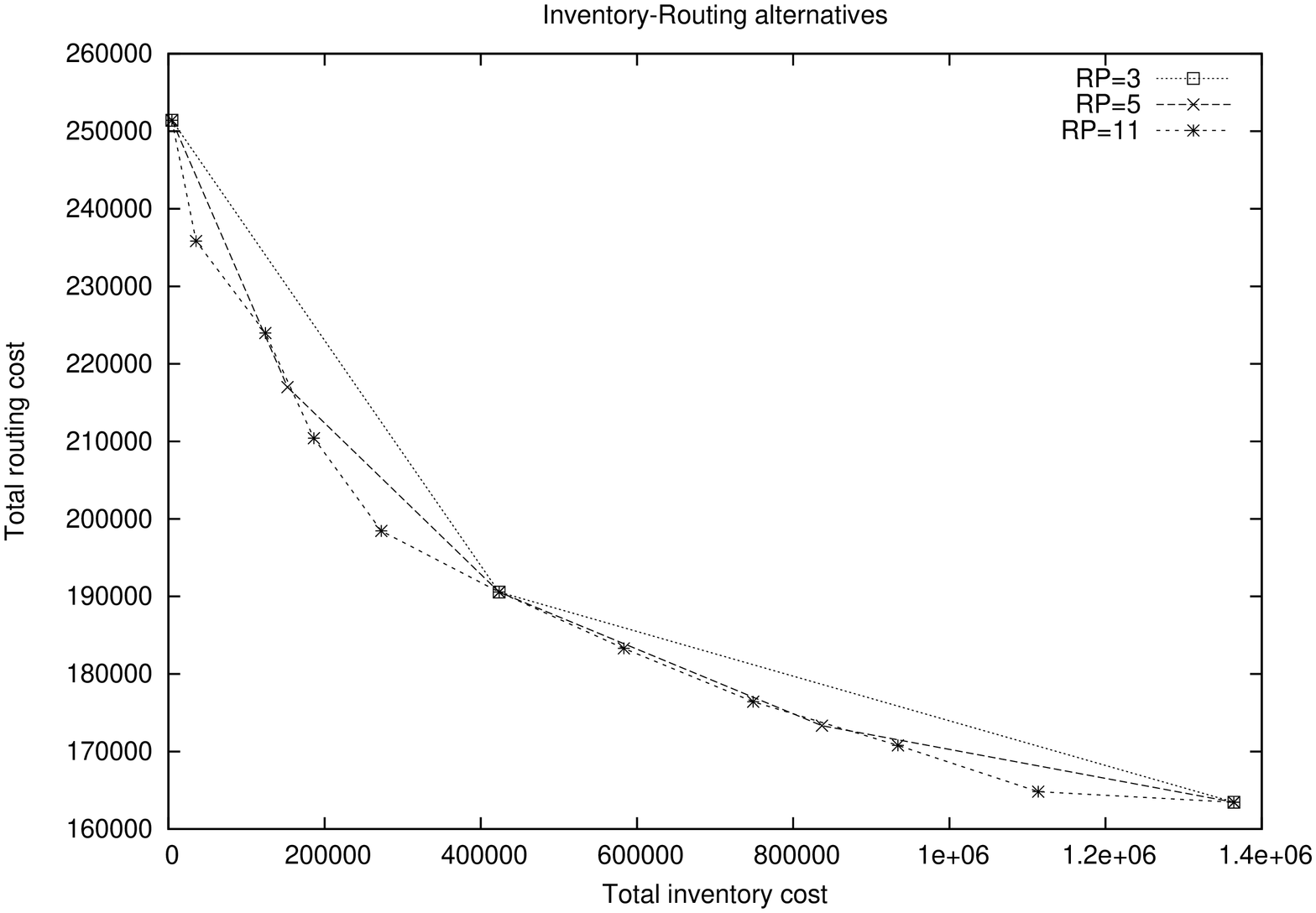}
    \includegraphics[angle=-90,width=.49\textwidth]{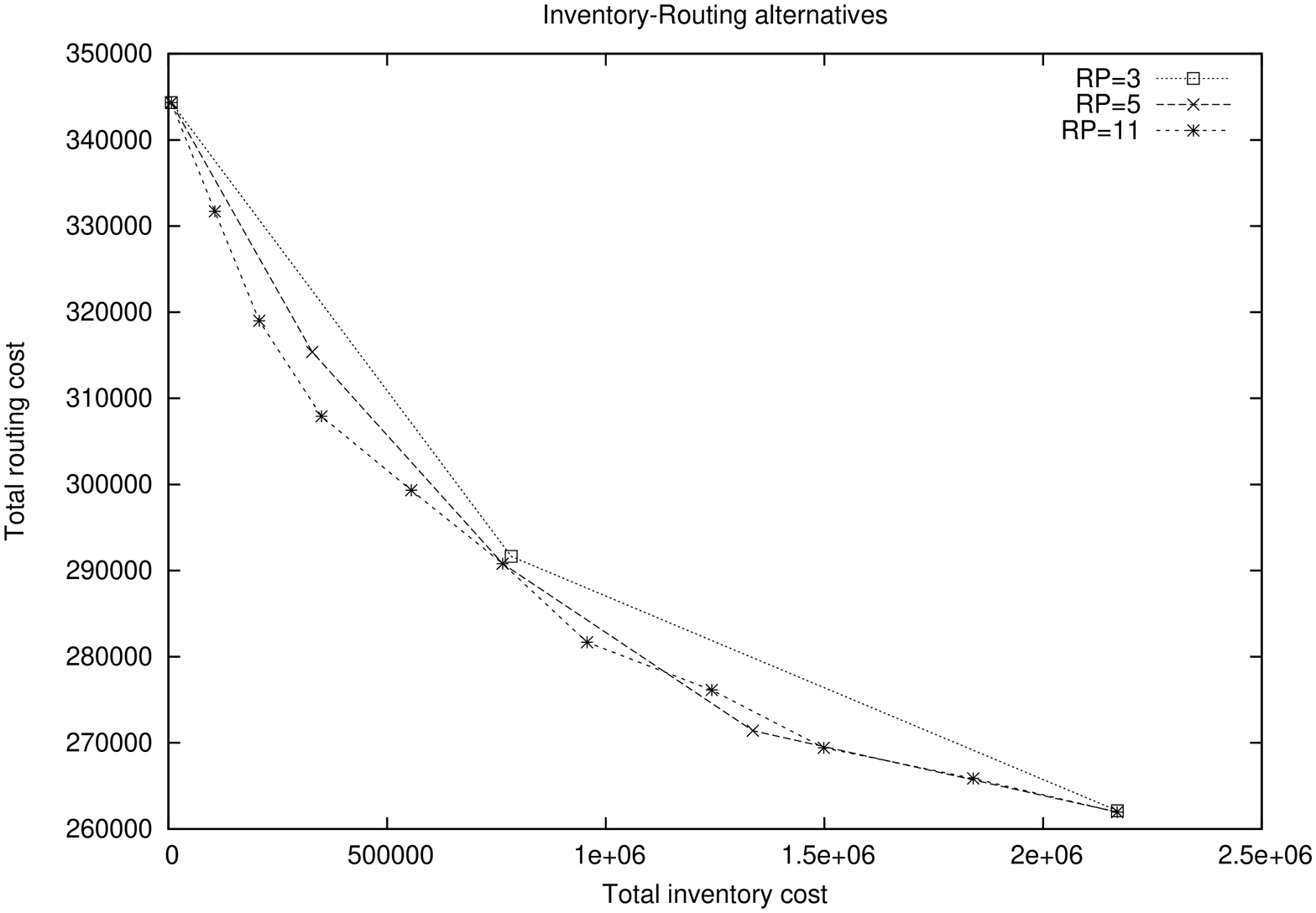}
  \caption{Results when testing a different number of reference points. Top left: instance {\tt GS-01-a.irp}, 50 customers, constant average demand (scenario a); top right: instance {\tt GS-02-a.irp}, 75 customers, constant average demand (scenario a); bottom left: instance {\tt GS-03-b.irp}, 100 customers, increasing average demand; bottom right: instance {\tt GS-04-c.irp}, 150 customers, average demand following a sinus function (scenario c).}
  \label{fig:plot}
\end{figure}

We have also closely analyzed the obtained results in outcome space, depending on the assumed number of reference points. More precisely, we investigated whether the results of a particular reference point setting lead to dominating results over other settings. Figure~\ref{fig:plot} reports four exemplary results, which act on behalf of the tested 12 cases.

The analysis shows that, for the problem at hand, even the outcomes obtained when assuming 3 reference points are helpful to the decision maker. Obviously, the approximation is improved by increasing the number of reference points at the cost of additional computations. In Figure~\ref{fig:plot}, this is shown by an increasing `convexity' of the Pareto-front.
Despite these refinements, the results of smaller reference point sets are however not dominated. Overall, and in combination with the above presented analysis of the running times, this implies that first approximations involving few reference points are not only comparably fast computable, but also acceptable with respect to the obtained quality.

\newpage

\section{\label{sec:conclusions}Conclusions}
The article presented a study on the biobjective IRP, a generalization of the classical IRP with its aggregated objective function. Challenging data sets comprising $T = 240$ periods have been investigated, which refer to applications in which optimization has to be done over a mid-term planning horizon.

A local search approach employing reference points has been presented, implemented, and tested on the given benchmark instances. The approach has been formulated in response to previous research~\cite{geiger:2011}, which clearly showed the difficulty of identifying all Pareto-optimal solution within reasonable time.

By means of reference points, we may modulate the computational effort and thus contribute to solving such large and computationally challenging problems. Besides, the experiments show that for the investigated problem, outcomes obtained by few reference points are not dominated by the ones of larger reference point sets. In conclusion, we may state that the proposed techniques presents a suitable approach for such biobjective Inventory Routing Problems, especially when being confronted with larger data sets.

Despite this progress, some open issues exist. On the one hand, the computing times are, in some cases, still large and should be improved as part of our future work. On the other hand, more effective local search strategies should be tested, i.\,e.\ better neighborhood structures that allow for a faster convergence.

\bibliography{mic-geiger-sevaux}
\bibliographystyle{plain}

\end{document}